\documentclass[runningheads]{llncs}

\usepackage{cite}
\usepackage{amsmath,amssymb,amsfonts}
\usepackage{algorithm}
\usepackage{algpseudocode}
\usepackage{diagbox} 
\usepackage{multirow}
\usepackage{multicol}
\usepackage{booktabs}
\usepackage{graphicx}
\usepackage{textcomp}
\usepackage{xcolor}
\usepackage{float}
\usepackage{threeparttable}
\usepackage{wrapfig}
\usepackage[hidelinks]{hyperref}
\usepackage{color,soul}
\usepackage{amsmath,amssymb}
\usepackage{tabularx}
\usepackage{circledsteps}
\usepackage{pifont}
\usepackage{bbding}

\begin{document}
\title{Disttack: Graph Adversarial Attacks Toward Distributed GNN Training}
\author{Yuxiang Zhang\inst{1,2} \and Xin Liu\inst{1,2} \and Meng Wu\inst{1} \and Wei Yan\inst{1,3} \and \\ Mingyu Yan\inst{1(}\Envelope\inst{)} \and Xiaochun Ye\inst{1} \and Dongrui Fan\inst{1}}
\authorrunning{Y. Zhang et al.}
\institute{SKLP, Institute of Computing Technology, Chinese Academy of Sciences
\email{\{zhangyuxiang22s, liuxin19g, wumeng, yanwei,\\ yanmingyu, yexiaochun, fandr\}@ict.ac.cn} \and University of Chinese Academy of Sciences $^3$ Zhongguancun Laboratory} 
%
\maketitle              
\begin{abstract}
Graph Neural Networks (GNNs) have emerged as potent models for graph learning. Distributing the training process across multiple computing nodes is the most promising solution to address the challenges of ever-growing real-world graphs.
However, current adversarial attack methods on GNNs neglect the characteristics and applications of the distributed scenario, leading to suboptimal performance and inefficiency in attacking distributed GNN training.

In this study, we introduce Disttack, the first framework of adversarial attacks for distributed GNN training that leverages the characteristics of frequent gradient updates in a distributed system. Specifically, Disttack corrupts distributed GNN training by injecting adversarial attacks into one single computing node. The attacked subgraphs are precisely perturbed to induce an abnormal gradient ascent in backpropagation, disrupting gradient synchronization between computing nodes and thus leading to a significant performance decline of the trained GNN. We evaluate Disttack on four large real-world graphs by attacking five widely adopted GNNs. Compared with the state-of-the-art attack method, experimental results demonstrate that Disttack amplifies the model accuracy degradation by 2.75$\times$ and achieves speedup by 17.33$\times$ on average while maintaining unnoticeability.

\keywords{Graph Neural Network  \and Distributed Training \and Adversarial Attack.}
\end{abstract}

\section{Introduction}
Recently, Graph Neural Networks (GNNs) \cite{gcn} have gained considerable attention as powerful models for learning graph data \cite{gnn1} in various tasks. With the explosion of information nowadays, the scale of real-world graph data is proliferating, which goes beyond training GNN on a single computing node. The most effective approach to mitigate inefficiency is distributed training, which distributes the training workload across multiple computing nodes. For example, NeuGraph \cite{distgnn1} can handle large graphs with millions of nodes and edges while achieving 16 $\sim$ 47 $\times$ speedups than training GNN on a single computing node.

\begin{figure*}[tb]
\centering
\includegraphics[width=0.8\textwidth]{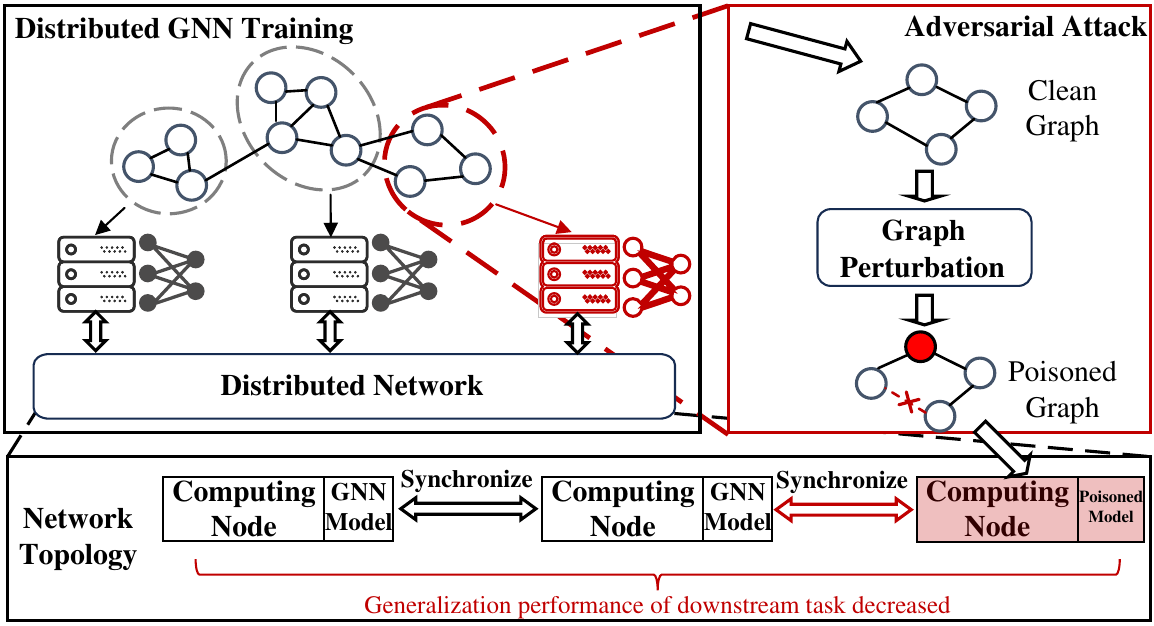}
\caption{Graph adversarial attack in distributed GNN training scenario.}
\label{fig:5}
\end{figure*}

However, distributed GNNs are susceptible to adversarial attacks involving the graph's structure and feature perturbations. Communication complexity among multiple computing nodes amplifies security challenges in distributed scenarios, as depicted in Fig. \ref{fig:5}. Unlike GNNs on a single computing node, distributed GNNs require synchronization between multiple computational nodes to update model parameters uniformly, as highlighted in previous research \cite{ley}. Adversaries are common, and false data can be easily injected into a distributed system. For example, fraudsters perturb one cloud node to affect the entire cloud system in cloud service \cite{seccloud}. Exploring the vulnerability of distributed GNNs under attack and revealing attack behaviors are crucial to improving the robustness of this commonly adopted paradigm due to the ubiquitous and destructive of adversarial attacks.

Existing adversarial attacks for GNNs \cite{patk1, patk2, sga} have overlooked the characteristics of distributed training, such as device communication, which leads to ineffective attacks on distributed GNN training. Meeting their significant computational and memory requirements also presents challenges for distributed systems. For example, the pioneer adversarial attack for graph data \cite{patk1} has failed to conduct attacks on large datasets. This gap highlights the critical need for research into attack methods specifically designed for distributed GNNs that can scale effectively without sacrificing attack efficiency. 



Therefore, We propose \textbf{Disttack}\footnote[1]{Open-sourced at \url{https://github.com/zhangyxrepo/Disttack}}, the first adversarial attack framework tailored for distributed GNN training to our knowledge. We exploit the vulnerabilities exposed during gradient synchronization, i.e., a unique execution phase divergent from single computing node GNN training. To enhance Disttack's time and space efficiency, we sample the subgraph consisting of $1$-hop neighbors of target nodes and target specific nodes and edges with the most significant impact within the subgraph. We then amplify gradient ascent by perturbing these identified node features and edges. The accumulation of these effects disrupts gradient synchronization and global training results. Extensive experiments show that Disttack outperforms existing state-of-the-art adversarial attack methods, causing more significant degradation of the GNN model's performance. Disttack brings up to 2.75 $\times$ and 2.62 $\times$ performance improvement compared to SGA \cite{sga} attack GAT \cite{gat} \ on Arxiv and Metattack \cite{patk2} attack GIN \cite{gin} on Reddit, respectively. Since maintaining unnoticeable is influential for adversarial attacks, we refine homophily distribution based on the existing method \cite{unnotice} to measure the unnoticeability of attacks and prove that Disttack is more stealthy than other baselines. Disttack also shows significant efficiency gains, up to 89.64 $\times$ faster on Reddit-SV and 17.33 $\times$ faster on Reddit than SGA and Mettack, respectively.


The main contributions of this paper can be summarized as follows:
\begin{itemize} 
    \item We propose a novel adversarial attack algorithm that combines the gradient-based attack approach with gradient synchronization during distributed GNN training. Moreover, we have developed it and ensured its unnoticeability for enhanced stealth.
    
    \item We introduce \textbf{Disttack}, a novel framework to inject adversarial attacks to distributed GNN training, which generates subtle yet effective perturbations to decrease the distributed GNN model's generalization performance. It presents a new design point by bridging the gap between adversarial attacks and distributed GNN training for the first time.
    
    \item We conduct extensive experiments for five GNN models across four real-world graph datasets. Results suggest that Disttack remains unnoticeable while offering considerable improvements in performance and efficiency compared with existing state-of-the-art attack methods.
\end{itemize}
    



\section{Related Work}


\noindent \textbf{Adversarial Attack Toward GNNs.} In the GNN-related domain, adversarial attacks are proposed to worsen the performance of downstream tasks of the GNN model. Our work related to two categories of previous adversarial attack approaches on the graph: poisoning attack and gradient-based attack \cite{adv0}. Poisoning attacks \cite{patk1} manipulate the victim model's training dataset, and this process is called poisoning. The objective of these attacks is to degrade the model's performance by injecting adversarial examples into the training dataset. Gradient-based attacks \cite{sga} leverage gradient information in generating adversarial samples. It often involves acquiring or approximating gradient information to identify features critical to the model's performance.

\noindent \textbf{Distributed GNN Training.} Distributed training is a popular solution to speed up GNN training by adding more computing nodes to the distributed system with parallel execution strategies. Typically, distributed GNN training methods can be divided into two categories \cite{ley}, i.e., distributed full-batch training \cite{neugraph} and distributed mini-batch training \cite{distgnn1}. The difference between them is that the former distributes different partitions of graphs to computing nodes, whereas the latter processes several mini-batches simultaneously to parallelize the training process. We focuses on mini-batch training since it has become mainstream due to less communication volume and memory consumption \cite{ley, distsurvey}.

Most existing attack methods \cite{patk1, patk2, sga} have been deployed on a single computing node and small-scale datasets. These approaches must be revised in today's escalating distributed system scale, which demands the full use of distributed GNNs to deal with graph tasks efficiently. Realizing this trend, we dedicate ourselves to formulating an adversarial attack framework tailored for distributed GNN training. It aims to scale attack strategies to address the challenges posed by modern data processing requirements.
\section{Methodology}
In this section, we first present an essential preliminary of our methodology. Subsequently, we elucidate the specifics of our approach, highlighting its innovative elements and the mechanics of its implementation.

\subsection{Preliminary of Attacks in Distributed GNN Training}
Consider a graph \( G \) with node features \( X \) and labels \( Y \). We define \( G = (V, E, X) \) as an undirected graph, where \( V = \{v_1, v_2, \ldots, v_N\} \) denotes set of \( N \) nodes, and \( E \subseteq V \times V \) denotes the set of edges. An adjacency matrix \( A \) represents the graph's spatial structure, indicating the nodes' topological relationship. We take a vanilla Graph Convolutional Network (GCN) \cite{gcn} as an exemplar and formalize the procedures of graph learning as follows:
\begin{equation}\label{eq-gcn}
H^{(l+1)} = \sigma\left(\tilde{D}^{-\frac{1}{2}} \tilde{A} \tilde{D}^{-\frac{1}{2}} H^{(l)} W^{(l)}\right),
\end{equation}
where $\tilde{A}$ represents the adjacency matrix includes self-loop, and $\tilde{D}$ is the degree matrix of $\tilde{A}$. Here, $H^{(l)}$ and $W^{(l)}$ denote the hidden features and the trainable weights at the $l^{th}$ layer. The initial feature is $H^{(0)} = X$. The function $\sigma(\cdot)$ signifies a non-linear function, such as the ReLU.

Transitioning from this to distributed GNN training, the synchronization of gradients across computing nodes is required to maintain the consistency of the model's parameters. The process can be formulated as:
\begin{equation} \label{dist-weight}
{W}_{i+1}={W}_i+\sum_{j=1}^n \nabla g_{i, j} ,
\end{equation}
where $W_i$ denotes the model's weight in the $i$ th round of computation, $\nabla g_{i, j}$ is the gradient generated in the backward propagation by computing node $j$ in the $i$ th round of computation, and $n$ is the number of the computing nodes.

Another essential formula is the loss function. As argued in the previous section, the loss functions of mainstream distributed GNN training methods can be expressed as follows:
\begin{equation} \label{dist-loss}
    \mathcal{L}=\frac{1}{|{V}_s|} \sum_{v_i \in {V}_s} \nabla l(y_i, z_i) ,
\end{equation}
where $\mathcal{L}$ is the loss computed over the iteration, ${V}_s$ is randomly sampled from the training dataset in each training iteration, $\nabla l(\cdot)$ signifies the loss function of the model, $y_i$ denotes the actual label of the node $v_i$, and $z_i$ to the predicted output of the GNN model for the input node $v_i$.

\begin{figure*}[tb]
\centering
\includegraphics[width=0.9\textwidth]{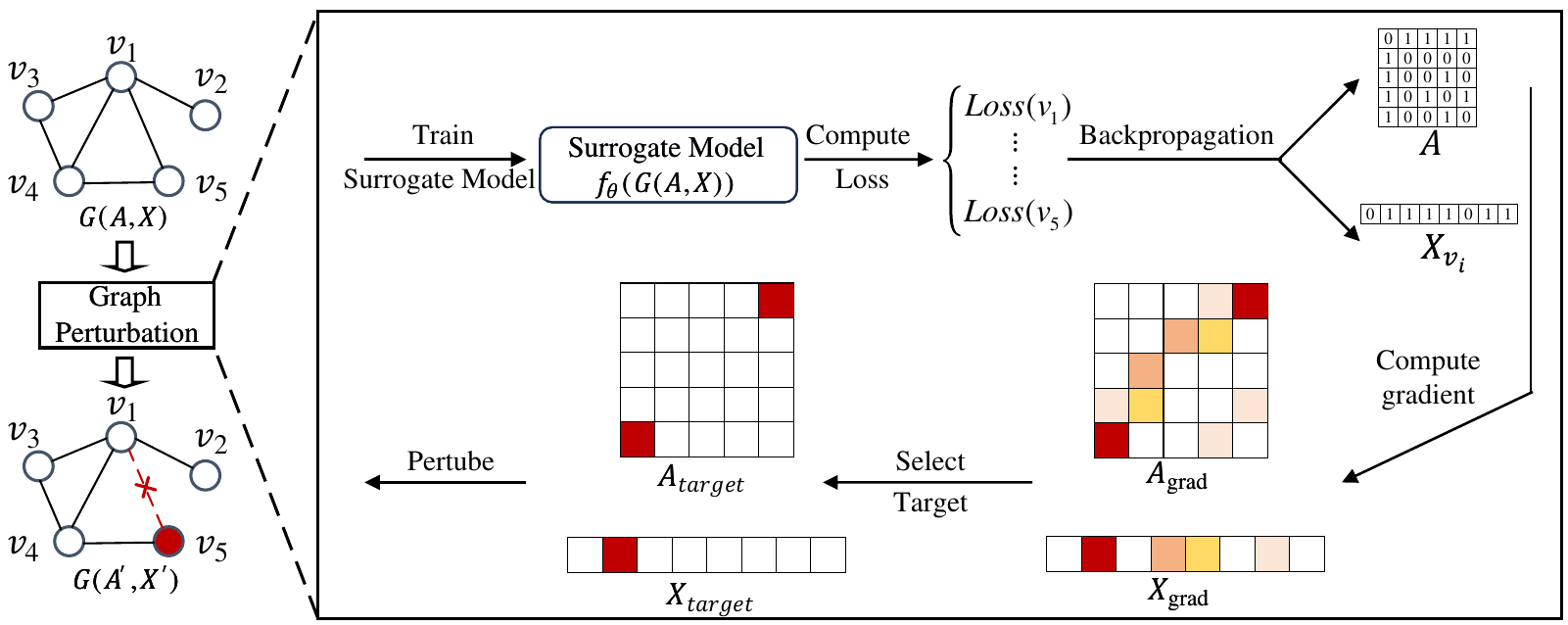}
\caption{Illustration of the example process of generating graph perturbation.}
\label{fig:1}
\end{figure*}

After detailing the preliminary in distributed GNN training, we introduce common adversarial attacks toward GNN. Notably, these methods significantly impact responses to gradient-based attacks, a prevalent technique for introducing graph perturbations. Fig. \ref{fig:1} provides an overview of the typical GNN adversarial attack process. As Fig. \ref{fig:1} illustrates, the attack strategy involves training a surrogate model, expressed as $f_{\theta}(G(A, X))$, with the given graph $G$. Given the discrete nature of the graph, directly applying the gradient to the adjacency matrix is not feasible. It can be converted to an optimal problem along with the gradients for the graph structure and node features as following equations:
\begin{equation}\label{eq-theta}
\begin{gathered}
\theta^* = \arg \min \mathcal{L}_{train}\left(f_\theta(G(A,X)), Y\right), \\
A_{grad} = \nabla_A \mathcal{L}_{atk}\left(f_{\theta^*}(G(A,X))\right), \\
X_{grad} = \nabla_X \mathcal{L}_{atk}\left(f_{\theta^*}(G(A,X))\right),
\end{gathered}
\end{equation}
where $\theta^*$ is the set of optimized parameters that minimize the training loss $\mathcal{L}_{train}$. The gradients $A_{grad}$ and $X_{grad}$ are of the adversarial loss $\mathcal{L}_{atk}$ concerning the adjacency matrix $A$ and the node features $X$. These gradients guide the selection of perturbations that are likely most effective in an adversarial attack \cite{sga}. The attack loss function $\mathcal{L}_{atk}$ is defined as:

\begin{equation}\label{eq-atkloss}
\mathcal{L}_{atk} = -\sum_{i=1}^{V} \mathcal{L}\left(f_{\theta^*}(G), y_i\right),
\end{equation}
where $V$ is the set of nodes targeted under attack, and $f_{\theta^*}$ symbolizes the optimized surrogate model. In edge-targeted attacks, $A_{i,j}=1$ means nodes $v_i$ and $v_j$ are connected. A negative gradient, $A_{grad_{i,j}} \leq 0$, implies that removing edge $E_{i,j}$ may harm the victim models.

Similarly, for node feature perturbation attacks, the feature vector of a node $v_i$ is denoted as $X_i$. If the gradient for the $j$-th feature dimension $X_{grad_{i,j}} \leq 0$, altering this feature could negatively affect the victim model. To generate perturbed adjacency matrix $A^{\prime}$ and node feature $X^{\prime}$ over several iterations can be formulated as:
\begin{equation}\label{eq-adjgrad}
\begin{gathered}
A_{\text{iter}}^{\prime}, X_{\text{iter}}^{\prime}=\phi\left(\nabla_A \mathcal{L}_{atk}\left(f_{\theta^*}\left(G_{\text{iter} - 1}(A, X)\right)\right), A_{\text{iter}-1}^{\prime}, X_{\text{iter}-1}^{\prime}\right)) 
\end{gathered}
\end{equation}
These equations indicate that the choice of perturbation is determined by the attack loss $\mathcal{L}_{atk}$, the attack strategy $\phi$, and the trained surrogate model $f_{\theta^*}$.


\subsection{Attack Methodologies from Edge and Node Perspectives}
Building on previous work \cite{adv0}, poisoning attacks can be mathematically formulated as a problem:
\begin{equation}
\begin{aligned}
&\min _{G^\prime \in \Phi(G^\prime)} \mathcal{L}_{\text{atk}}\left(f_{\theta^*}(G^\prime)\right) \\
&\text{s.t. } \theta^*=\underset{\theta}{\arg \min } \phantom{1} \mathcal{L}_{\text{train}}\left(f_\theta(G^\prime)\right) ,
\end{aligned}
\end{equation}
where $\Phi(G^\prime)$ signifies the set of feasible perturbations on an original graph $G$, wherein $G^\prime$ is the perturbed graph. Prior work \cite{sga} has explored the attack by sampling the $k$-hop subgraph of the training nodes. In theory, this method connects the non-existent edges to reinforce the attack. the sampling process in isolation imposes a time order of $O(d^k)$, with $d$ denoting the average node degree. However, such demands become impractical in distributed GNN training, where large datasets are common \cite{ogbn}.

To address this issue, we adopt a simplified strategy that targets the first-order neighbors of the nodes. Utilizing the gradients of the training loss $\mathcal{L}_{\text{train}}$, we can efficiently identify and remove critical edges to maximize the loss and weaken the GNN's performance. Based on the findings in \cite{patk2}, the method is summarized by the formula:
\begin{equation}\label{eq-subgrad}
\nabla_{G^{(sub)}} = w_A \cdot \nabla_{A^{(sub)}} {\mathcal{L}}_{\text{train}}^{(sub)} + w_X \cdot \nabla_{X^{(sub)}} {\mathcal{L}}_{\text{train}}^{(sub)},
\end{equation}
where $w_A$ and $w_X$ indicate the weight of the structure and feature changes in the gradient calculation.

In edge perturbations, we introduce a score matrix $S$. Each element $S_{i,j}$ quantifies the expected increase in the attack loss $\mathcal{L}_{\text{atk}}$, resulting from the removal of edge $E_{i,j}$. To calculate these scores, we examine $\mathcal{L}_{\text{atk}}$ in response to the adjacency matrix, $A^{(sub)}$, of the subgraph $G^{(sub)}$. Drawing on from previous work \cite{patk2}, we enhance our scoring technique by initialing the gradient $\nabla_{A^{(sub)}} \mathcal{L}_{\text{atk}}$, which exposes the degree to which the adversarial loss is affected by each edge in $G^{(sub)}$. We then reform the score matrix as follows:
\begin{equation}
\begin{aligned}
S^{(sub)}_{i,j} = A^{(\text{sub})} &\odot \left(G^{(\text{sub})} + \lambda_{comm} \cdot C^{(\text{sub})}\right), \\
\text{s.t.} \quad C^{(\text{sub})}_{i,j} &= 
\begin{cases}
\phantom{-} 1, & \text{if } {CN}_i \neq {CN}_j \\
-1, & \text{otherwise.}
\end{cases}
\end{aligned}
\end{equation}
where $\odot$ symbolizes element-wise product,  $\lambda_{\text{comm}}$ is the weight of loss $C^{(\text{sub})}$ that arises from nodes processed by disparate computing nodes. Prior research \cite{distgnn1} indicates that edges linking nodes across different computing nodes can reduce accuracy in GNN training relative to single-node setups. Our approach exploits it by considering whether the edge connects nodes on the same computing node $CN$ while computing the score matrix $S$.

To point the node features that will enhance the adversarial loss $\mathcal{L}_{\text{atk}}$, we present the gradient of $\mathcal{L}_{\text{atk}}$ concerning the node features of the subgraph as $\nabla_{X^{(sub)}} \mathcal{L}_{\text{atk}}$. Our objective is to identify one element of the feature vector that, when perturbed, will increase $\mathcal{L}_{\text{atk}}$. This vector is obtained from the gradient of $\mathcal{L}_{\text{atk}}$ with respect to the features of a node $i$, represented as $\nabla_{X_i} \mathcal{L}_{\text{atk}}$. By disrupting the node features to the opposite direction of gradient descent, we aim to realize the greatest decrease in $\mathcal{L}_{\text{atk}}$, which can be formulated as:

\begin{equation}
X_{i+1}' =  X_i \cdot (1 - 2 \cdot \text{sgn}(\nabla_{X_i} \mathcal{L}_{\text{atk}})),
\end{equation}
where $\text{sgn}(\nabla_{X_i} \mathcal{L}_{\text{atk}})$ gives the direction of reversing the features to maximize the decrease in $\mathcal{L}_{\text{atk}}$.

Being stealthy in adversarial attacks is significant for attackers. Inspired by prior work \cite{unnotice}, we propose an improved metric, homophily distribution, to measure the unnoticeability of attack. We refine this metric to be more concise and reasonable by simplifying the computation and amplifying the influence of suspicious nodes. The homophily $h_i$ of a node $i$ is quantified by the similarity between the node's features and those of its neighbors, defined as:

\begin{equation}\label{eq-hi}
h_i=||\left(a_i, X_i\right)||_2, \quad a_i=\sum_{j \in \mathcal{N}(i)} \frac{\sqrt{d_j}}{\sqrt{d_i}} X_j ,
\end{equation}
where $a_i$ denotes the aggregated features of node $i$'s neighbors, and $d_i$ represents the degree of node $i$. In response, we refine our adversarial loss function to integrate this stealth measure, reformulating our objective as:
\begin{equation}\label{eq-hg}
\begin{gathered}
\min_{G^\prime \in \Phi(G)} \mathcal{L}_{\text{atk}}\left(f_{\theta}(G^\prime)\right) + \lambda_{homo} \left|\mathcal{M}({\mathcal{H}_G} - {\mathcal{H}_{G^\prime}})\right| \\
\text{s.t.} \phantom{-} \theta^* = \underset{\theta}{\arg \min } \mathcal{L}_{\text{train}}\left(f_{\theta}(G^\prime)\right) .
\end{gathered}
\end{equation}

\begin{algorithm}[bt]
\caption{Poisoning Attack on Distributed GNN Training}
\label{alg:disttack}
\begin{algorithmic}[1]
\Require Graph $G(A, X)$, loss functions $\mathcal{L}_{\text{train}}$, $\mathcal{L}_{\text{atk}}$, weights $w_A$, $w_X$
\Ensure Perturbed graph $G'(A', X')$
\State Initialize $G' \gets G$
\For{each training iteration}
    \State Compute optimal parameters: $\theta^* \gets \arg \min_{\theta} \mathcal{L}_{\text{train}}(f_\theta(G'))$
    \State Compute weighted attack gradients: $ w_A \cdot\nabla_{A^{(sub)}} \mathcal{L}_{\text{atk}}$, $ w_X \cdot\nabla_{X^{(sub)}} \mathcal{L}_{\text{atk}}$
    \For{each edge $e_{i,j}$ in $A^{(sub)}$}
        \If{$A_{i,j} = 1$}
            \State Calculate edge score: $S^{(sub)}_{i,j} \gets A^{(sub)} \odot (G^{(sub)} + \lambda_{comm} \cdot C^{(sub)})$
            \State Remove edge $e_{i,j}$ based on score $S^{(sub)}_{i,j}$
        \EndIf
    \EndFor
    \For{each node feature $X_i$ in $X^{(sub)}$}
        \State Perturb feature vector: $X_i \gets X_i \cdot (1 - 2 \cdot \text{sgn}(\nabla_{X_i} \mathcal{L}_{\text{atk}}))$
    \EndFor
    \State Update perturbed subgraph: $G' \gets G^{(sub)}(A', X')$
\EndFor\\
\Return $G'$
\end{algorithmic}
\end{algorithm}

Here, $\lambda_{homo}$ is a hyperparameter that adjusts the weight of the homophily distribution difference in the loss function, and $\mathcal{M}(\cdot)$ is a measure of the disparity between distributions. $\mathcal{H}(\cdot)$ signifies the homophily distribution of a graph, encapsulating the proposed stealth criterion within our adversarial framework.
\begin{table}[bt]
  \centering
  \caption{Comparative Analysis of Time Complexity}
  \begin{tabular}{@{}r>{\centering\arraybackslash}m{6cm}@{}} 
    \toprule[1pt] 
    \textbf{Attack Method} & \textbf{Time Complexity} \\
    \midrule
    Metattack & $O((|A^{(sub)}| + M)\cdot (N\cdot d)^2$) \\
    SGA & $O((|A^{(sub)}| + M)\cdot (|A^{(sub)}| + M) \cdot d^k \cdot N$) \\
    Disttack & $O(N \cdot (|A^{(sub)}| \cdot d + M))$ \\
    \bottomrule[1pt] 
  \end{tabular}
  \label{tab:cost}
\end{table}

We present an overall Algorithm \ref{alg:disttack}, designed for conducting poisoning attacks on distributed GNN training. The algorithm modifies the input graph $G$ to produce a perturbed graph $G'$ to minimize adversarial loss $\mathcal{L}_{\text{atk}}$ while preserving the structure of the original graph. The algorithm operates iteratively during training by perturbing structure and node features.

Building upon the procedure outlined in Algorithm \ref{alg:disttack}, we have thoroughly analyzed the time complexity associated with our Disttack algorithm. Assuming that the sampled subgraph comprises $N$ nodes, each with dimension vectors of feature $M$, and $|A^{(sub)}|$ represents the number of edges within the sampled subgraph, the average degree of the node being $d$, the time complexity of Disttack can be expressed as $O(N \cdot (|A^{(sub)}| \cdot d + M))$. 

To better understand the efficiency of our method, we compare the time complexity of our method with two state-of-the-art methods, Metattack and SGA, in Table \ref{tab:cost}. According to analysis in Table \ref{tab:cost}, the time cost of Metattack and SGA will increase seriously when the graph becomes dense and large, which will be further detailed in Section \ref{exp-time}.

\subsection{Disttack: Framework Overview}

We focus on developing a novel attack framework to decrease the generalization performance of the distributed GNN model on node classification tasks. The primary objective is to degrade the quality learned by a distributed GNN model during training. Since attacking all computing nodes is unrealistic, one way to approach this is to perturb a single computing node maximally to impact the entire system due to synchronization in the distributed GNN training.

\begin{figure*}[tb]
\centering
 \includegraphics[width=\textwidth]{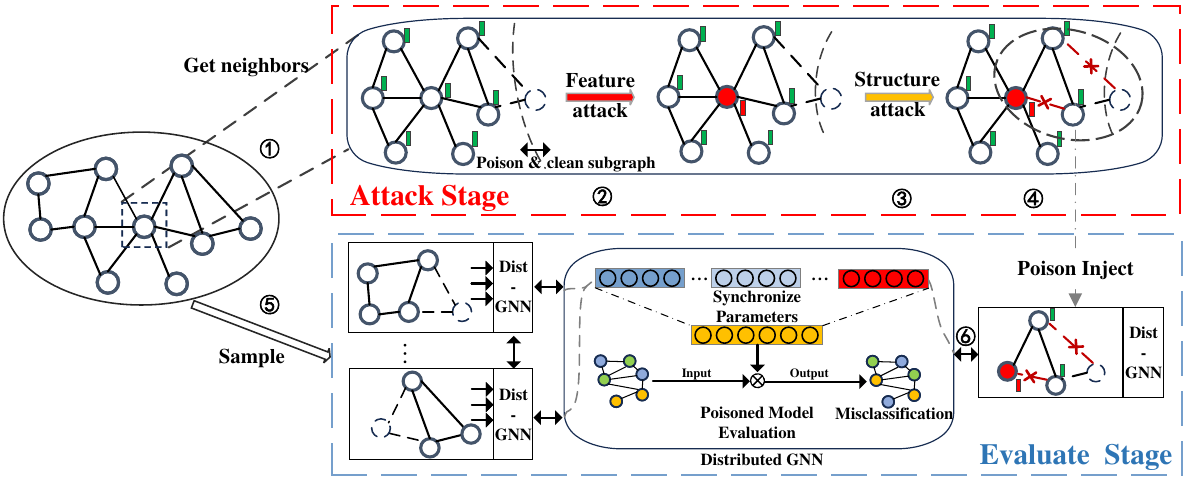}
\caption{The overall framework of the Disttack.}
\label{fig:2}
\end{figure*}

We first present an overview of our attack framework, Disttack, as shown in Fig. \ref{fig:2}. We can explain the process underlying Disttack's adversarial attack against distributed GNN concerning the numerical annotations. During the Attack Stage, Disttack \Circled{1} targets a node by sampling its $1$-hop neighbors, considering all connected nodes, including those on other computing nodes. Disttack then attacks the \Circled{2} node feature and \Circled{3} edges of samples. Following this, the adversarial samples created by Disttack are injected into the poisoned node to train the GNN model, as shown in \Circled{4}. Simultaneously, \Circled{5} the remaining computing nodes sample and train with clean data. 
During training, GNN models distributed on different computing nodes continuously synchronize their respective model parameters. \Circled{6} Disttack takes effect when the computing node injected poisoned training samples synchronizes its training parameters, including gradient and model weights, with other computing nodes. This poisoned synchronization process will influence the entire distributed GNN training system.

\begin{table}[b]
  \centering
  \caption{Dataset summary}
  \begin{tabular}{@{}r >{\centering\arraybackslash}p{2.2cm} >{\centering\arraybackslash}p{1.8cm} >{\centering\arraybackslash}p{2cm} cc@{}} 
    \toprule[1pt]
    \textbf{Dataset} & \textbf{Nodes} & \textbf{Edges} & \textbf{Avg Degree} & \textbf{Features} & \textbf{Classes} \\
    \midrule
    Flickr    & 89,250 & 225,270 & 2.52 & 50 & 7 \\
    Arxiv     & 169,343 & 2,315,598 & 13.67 & 128 &40 \\
    Reddit-SV & 232,965 & 23,213,838 & 99.65 & 602 & 41 \\
    Reddit    & 232,965 & 114,615,892 & 491.98 & 602 & 41 \\
    \bottomrule[1pt]
  \end{tabular}
 \label{tab:dataset}
\end{table}
\section{Experiment}



In this section, we conduct extensive experiments to demonstrate the effectiveness of Disttack and give a series of detailed analyses based on the results. By comparing the difference between the gradient norm of the poisoned computing node and the average, the experimental results confirm that Disttack outperforms its state-of-the-art counterparts while maintaining time efficiency and unnoticeability.

\subsection{Experimental Settings}\label{exp-setting}
\noindent \textbf{Datasets.}
To evaluate our method, we conduct experiments on datasets Flickr \cite{flickr}, Arxiv \cite{ogbn}, Reddit \cite{reddit}, and a sparse version of Reddit, denoted as Reddit-SV \cite{flickr}, which are detailed in Table \ref{tab:dataset}. Notably, the scalability of baselines is a significant limitation for choosing larger datasets, with many unable to handle even the smallest datasets we considered. For example, an estimation of the resources required by attack methods \cite{sga} on Reddit suggests a prohibitive 500 GB of memory, while our approach operates effectively at 12 GB.

\noindent \textbf{Baselines and Configuration.}
Disttack is evaluated on five widely-adopted GNN models, containing GCN \cite{gcn}, SGC \cite{sgc}, GIN\cite{gin}, GAT \cite{gat} and GraphSAGE \cite{flickr}. We compare Disttack with four baselines: \textbf{RA} perturb graph by randomly adding or removing edges; \noindent \textbf{DICE} \cite{dice} perturbs the graph by removing edges between nodes with the same label and adding edges between nodes with different labels; And two state-of-the-art attack method: \textbf{Metattack} \cite{patk2} pioneers the use of meta-gradients in adversarial attacks, which remains the leading global attack method in the field; \noindent \textbf{SGA} \cite{sga} is a simplified gradient-based attack framework that designs attacks based on gradient ascent. Specifically for \textbf{RA}, we use ten different seeds to alleviate the influence of randomness.

We are assuming that there are $N$ computing nodes in a distributed system. In our setup, each computing node sample from one of the $\frac{1}{N}$ share of the training dataset, respectively. Disttack and other baselines will perturb one of the shares. Baselines adhere to their reported configurations while ensuring that the number of perturbations aligns with that of Disttack. For fairness, we evaluate each experiment five times continuously and record the average value.

\noindent \textbf{Platforms.}
All experiments are conducted on a Linux server with a 32-core Intel Xeon Platinum 8350C CPU and four NVIDIA A100 SXM 80GB GPUs.
\vspace{-0.25cm}

\subsection{Analysis of Attacking Performance}\label{exp-performance} 
We demonstrate the effectiveness of Disttack through a comparison with different baselines, as shown in Table \ref{tab:exp1} and \ref{tab:exp2}. The best results are \textbf{bolded} while the second best are \underline{underlined}, and symbol $\textbf{-}$ indicates a failure attack due to run-time error. In general, Disttack consistently surpasses other baselines in all cases. The most significant performance improvement is recorded as 2.75$\times$, which occurs during comparison with the SGA attack GAT on Arxiv.

Focusing on the results for smaller dataset Flickr, the results in Table \ref{tab:exp1} indicate a decline in model performance, with GAT reporting the most pronounced drop, while GraphSAGE shows relative robustness. These results are consistent with findings in \cite{sga}, and explanations are rooted in model characteristics: (a) GAT's attention mechanism, which weights subgraph inputs, may exacerbate the effect of poisoned features. (b) The iterative nature of GraphSAGE's message aggregation could confer a better defense against structural perturbations. (c) Linear activation functions across all models except SGC permit the propagation of adversarial influence. Though simplified, the absence of hidden layers in the SGC may reduce its robustness.

In Table \ref{tab:exp2}, we extend our analysis to the Reddit-SV and Reddit, exploring the impact of Disttack in larger-scale datasets. The results consistently demonstrate state-of-the-art performance degradation across all GNN models under the influence of Disttack, as Table \ref{tab:exp1} has shown. Notably, two weaker baselines approached ineffectiveness due to the inherent complexity of the graph, which is characterized by high node degrees and edge density. Specifically, SGA failed to conduct the attack on Reddit due to runtime errors caused by memory overflow. In contrast, Disttack performed better than its counterparts.

\begin{table*}[tb]
  \centering
  \caption{Comparison of model accuracy on Flickr and Arxiv after attacking.}
  \resizebox{1\textwidth}{!}{%
    \begin{tabular}{@{}rcccccccccc@{}} 
    \toprule
    \multicolumn{1}{c}{Dataset} & \multicolumn{5}{c}{Flickr} & \multicolumn{5}{c}{Arxiv} \\
    \cmidrule(lr){2-11}
    \multicolumn{1}{c}{Backbone}& GCN & SGC & GAT & GIN & GraphSAGE & GCN & SGC & GAT & GIN & GraphSAGE \\
    \midrule
    Clean-Single & 0.5023 & 0.4871 & 0.5112 & 0.5015 & 0.5401 & 0.6875 & 0.6127 & 0.6547 & 0.6658 & 0.7231 \\
    Clean-Distributed & 0.5014 & 0.4779 & 0.5078 & 0.5004 & 0.5378 & 0.6872 & 0.6131 & 0.6546 & 0.6656 & 0.7228 \\
    \midrule
    RA & 0.4764 & 0.4435 & 0.4719 & 0.5056 & 0.5276 & 0.6410 & 0.6219 & 0.6634 & 0.6587 & 0.7186 \\
    DICE & 0.4821 & 0.4610 & 0.4677 & 0.4876 & 0.5102 & 0.6858 & 0.5779 & 0.6297 & 0.6423 & 0.6987 \\
    Metattack & 0.4924 & 0.4780 & \underline{0.4013} & 0.4798 & \underline{0.4931} & 0.6014 & \underline{0.2243} & 0.6211 & \underline{0.4314} & 0.6035 \\
    SGA & \underline{0.4771} & \underline{0.4513} & 0.4227 & \underline{0.4565} & 0.5050 & \underline{0.5130} & 0.2535 & \underline{0.6013} & 0.4342 & \underline{0.5897} \\
    \midrule
    \textbf{Disttack} & \textbf{0.4234} & \textbf{0.4193} & \textbf{0.3775} & \textbf{0.4412} & \textbf{0.4715} & \textbf{0.1713} & \textbf{0.1645} & \textbf{0.3331} & \textbf{0.4267} & \textbf{0.4435} \\
    \bottomrule
    \end{tabular}
  }
  \label{tab:exp1}
\end{table*}

\begin{table*}[tb]
  \centering
  \caption{Comparison of model accuracy on Reddit-SV and Reddit after attacking.}
  \resizebox{1\textwidth}{!}{%
  \begin{tabular}{@{}rcccccccccc@{}} 
    \toprule
    \multicolumn{1}{c}{Dataset} & \multicolumn{5}{c}{Reddit-SV} & \multicolumn{5}{c}{Reddit} \\
    \cmidrule(lr){2-11}
    \multicolumn{1}{c}{Backbone}& GCN & SGC & GAT & GIN & GraphSAGE & GCN & SGC & GAT & GIN & GraphSAGE \\
    \midrule
    Clean-Single & 0.5152 & 0.6658 & 0.7454 & 0.5532 & 0.6406 & 0.8958 & 0.9311 & 0.9520 & 0.7270 & 0.9301 \\
    Clean-Distributed & 0.5152 & 0.6431 & 0.7406 & 0.5529 & 0.6411 & 0.8804 & 0.9300 & 0.9401 & 0.7270 & 0.9281 \\
    \midrule
    RA & 0.5176 & 0.6102 & 0.7342 & 0.5011 & 0.6241 & 0.8707 & 0.9176 & 0.9501 & 0.7114 & 0.9117 \\
    DICE & 0.5288 & 0.6316 & 0.7101 & 0.5413 & 0.6337 & 0.8779 & 0.8976 & 0.9341 & 0.7129 & 0.9045 \\
    Metattack & 0.5092 & 0.6003 & 0.6239 & \underline{0.4475} & \underline{0.4730} & \underline{0.8653} & \underline{0.8893} & \underline{0.8714} & \underline{0.6435} & \underline{0.8753} \\
    SGA & \underline{0.4991} & \underline{0.5661} & \underline{0.5775} & 0.4607 & 0.6355 & - & - & - & - & - \\
    \midrule
    \textbf{Disttack} & \textbf{0.4404} & \textbf{0.4978} & \textbf{0.5547} & \textbf{0.3658} & \textbf{0.4631} & \textbf{0.6773} & \textbf{0.8801} & \textbf{0.8330} & \textbf{0.5920} & \textbf{0.7832} \\
    \bottomrule
  \end{tabular}
  }
  \label{tab:exp2}
\end{table*}

\subsection{Analysis of Gradients in Distributed Training.} 
\begin{figure*}[bt]
\centering
\includegraphics[width=0.94\textwidth]{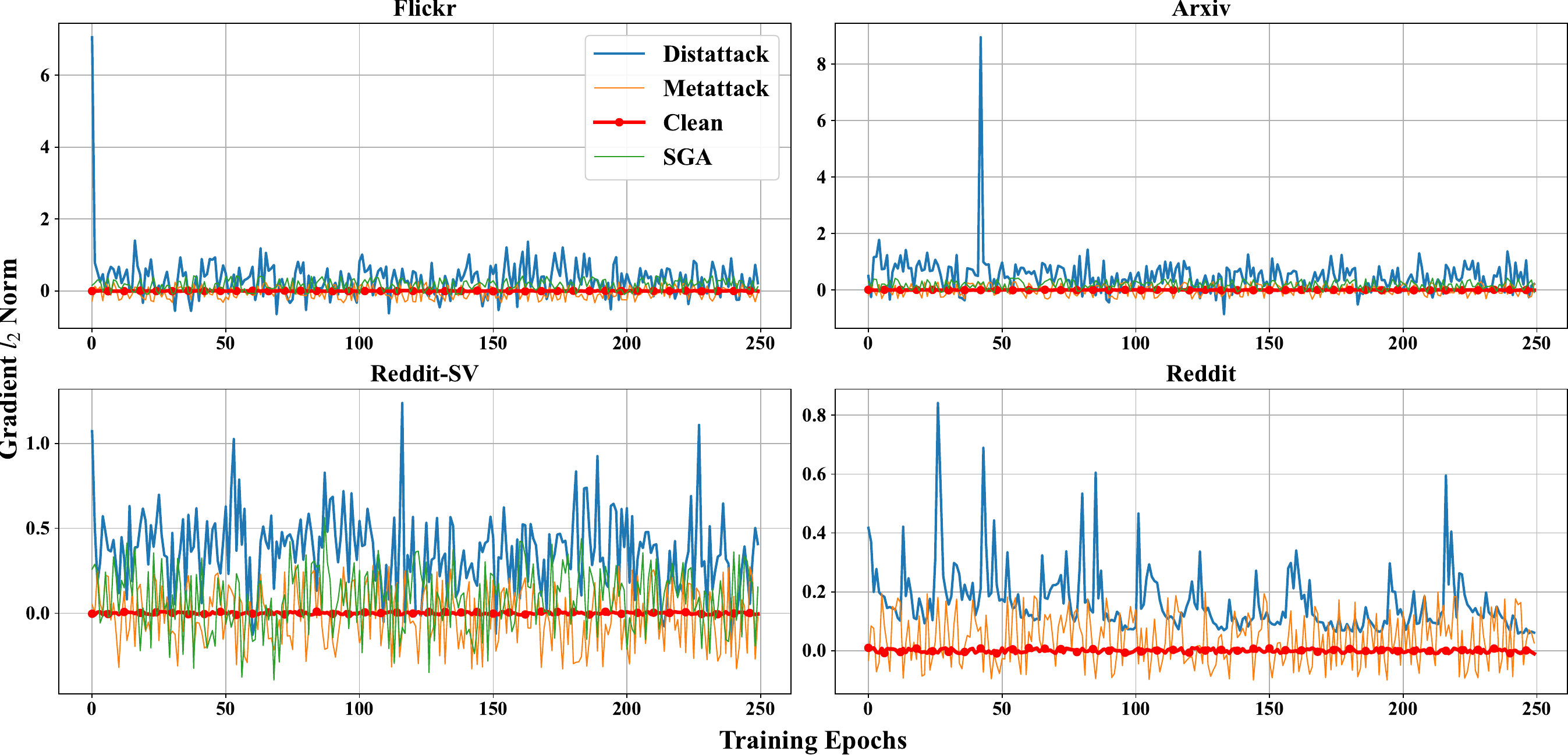}
\caption{Gradient $l_2$ norm variations of a 2-layer GCN under different attacks during distributed training.}
\label{fig:3}
\end{figure*}
Our method, Disttack, is designed to leverage gradient ascent in attacking distributed GNNs. The synchronization of gradients and model parameters among computing nodes is the crucial difference between distributed and single node GNN training. Different computing nodes' gradient $l_2$ norms will typically converge towards their mean value \cite{disttrain2}. Disttack precisely perturbed graphs to generate adversarial samples induces an abnormal gradient ascent in backpropagation, disrupting gradient synchronization between computing nodes and thus leading to a significant performance decline of the model. 

Fig. \ref{fig:3} illustrates the difference between the gradient $l_2$ norms of the model on the attacked computing node by different methods and the average gradient $l_2$ norms of those on clean computing nodes of 250 training epochs. As Fig.\ref{fig:3} illustrates, Disttack causes an abnormal increase in the gradient $l_2$ norm during distributed GNN training, and this increase caused by Disttack is more pronounced than that caused by other attacks. Specifically, gradient $l_2$ norms will show a pulse-like increase almost every ten training epochs on the Reddit-SV dataset due to Disttack. In contrast, existing methods do not fully leverage the characteristics of distributed GNN training. For example, SGA and Metattack hardly caused gradient abnormalities on the Arxiv or Flickr datasets. Notably, the impact caused by Distack will not be eliminated as the model training continues and parameters are updated, revealing the effectiveness of Disttack.

\vspace{-0.25cm}

\subsection{Analysis of Unnoticeability}\label{exp-unnotice}
\begin{figure*}[bt]
\centering
\includegraphics[width=0.94\textwidth]{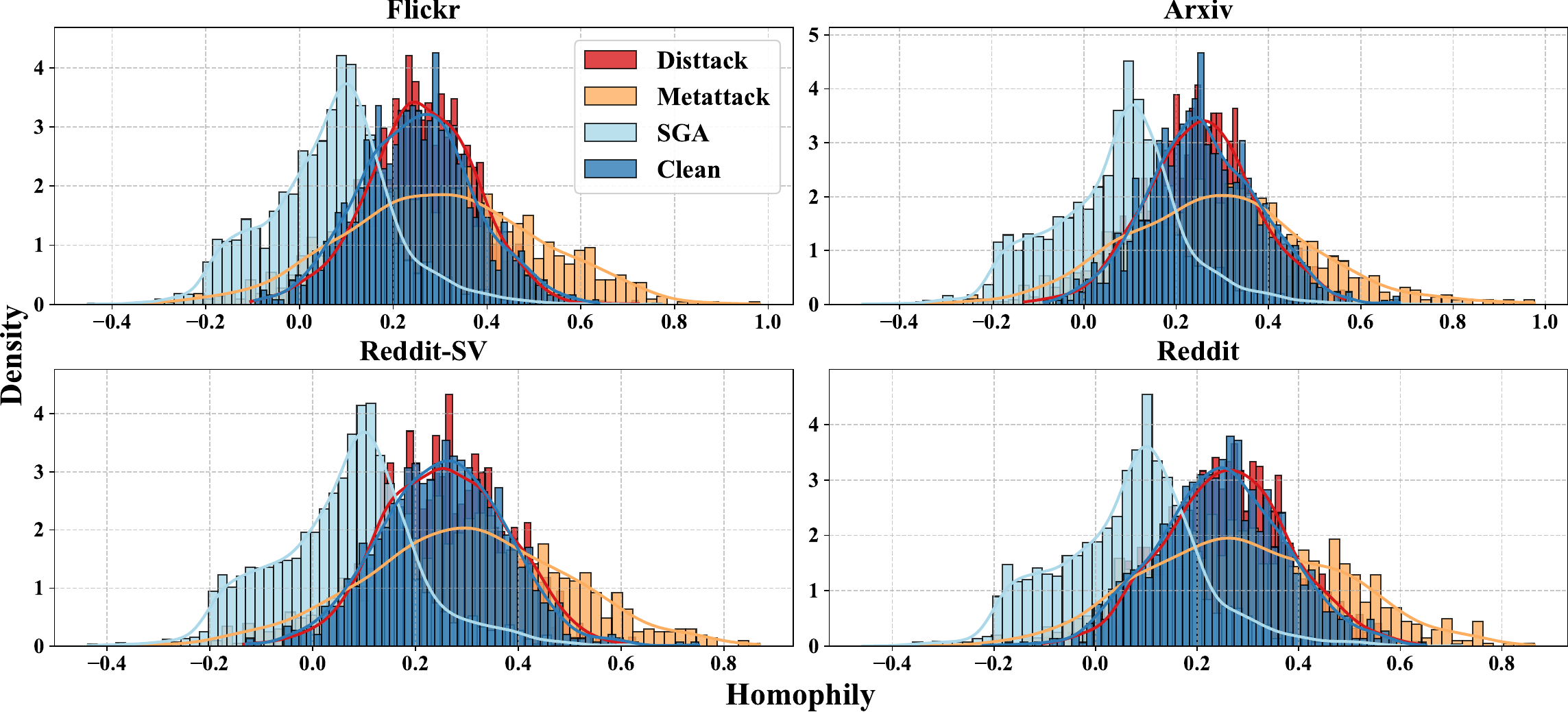}
\caption{Homophily changes of four datasets after attacking.}
\label{fig:6}
\end{figure*}
Building on the method to compute the homophily distribution described earlier \cite{unnotice}, we intuitively posit that the homophily distribution can serve as a metric for the imperceptibility of adversarial attacks in graphs. In a clean graph, this distribution is discernible. A deviation from this pattern compromises the attack's stealth, making the adversarial nodes and edges easily detectable and removable by database administrators or defenders against adversarial attacks.

Disttack adeptly maintains the homophily of the poisoned graph, aligning it closely with the clean graph to cover the attack, as shown in Fig. \ref{fig:6}, which is a result of adding homophily distribution into our optimized objective. In contrast, Metattack and SGA show a significant shift in homophily distribution. Although Metattack set budget constraints on the attack, maximizing the rank of the adjacency matrix caused the graph to be perturbed significantly. SGA maximally perturbs node features and edges, making it easy to detect.

\subsection{Analysis of Time Cost}\label{exp-time}
\begin{figure*}[tb]
\centering
\includegraphics[width=\textwidth]{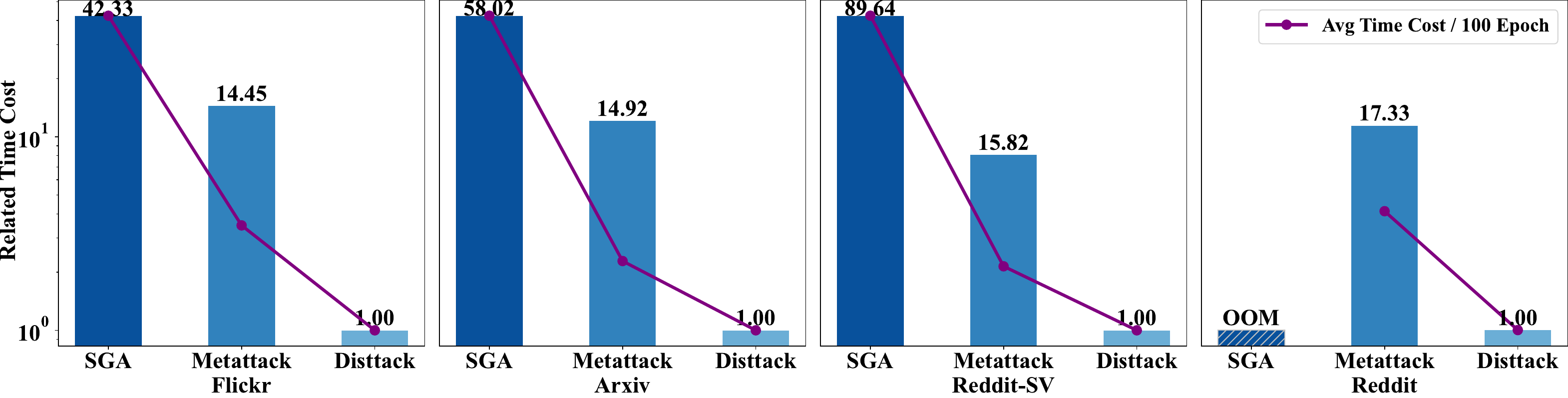}
\caption{Related time cost of attack methods. Note that the result for SGA on Reddit is unavailable due to a run-time error.}
\label{fig:4}
\end{figure*}
In evaluation, we measured the running time of Metattack and SGA relative to that of Disttack, as illustrated in Fig. \ref{fig:4}. For fairness, we also plot the average time required for different methods to run 100 epochs. Disttack still guarantees its efficiency in all cases. For example, compared to Metattack, Disttack achieved almost 10 $\times$ speedup by running 100 epochs to perturb Reddit. Moreover, we observed that this efficiency is more pronounced on dense graphs. For instance, Disttack shows up to 89.64 $\times$ speedup over SGA to perturb Reddit-SV. SGA shows poor time efficiency as it constantly computes a larger subgraph's gradients. Metattack treats the whole subgraph structure as a hyperparameter, which requires more extended execution on dense graphs. These results are approximately consistent with the analysis in Table \ref{tab:cost}, underscoring that Disttack is more time-efficient in attacking distributed GNN training.

\section{Conclusion}
In this work, we study the effects of adversarial attacks in distributed GNN training for the first time. Accordingly, we propose Disttack, which leverages the gradient synchronization properties of distributed GNN training to perform significant but subtle adversarial attacks. Disttack aims to poison the training dataset of one computing node and decrease the generalization performance of the whole distributed GNN model. Experiments confirm that Disttack outperforms other state-of-the-art attack methods, amplifying model accuracy degradation up to 2.75$\times$. Furthermore, we conducted a quantitative analysis and verified that Disttack achieves a 17.33 $\times$ speedup over a state-of-the-art baseline while maintaining an unnoticeable attack.


\section*{Acknowledgement} This work was supported by National Key Research and Development Program (Grant No. 2023YFB4502305), the National Natural Science Foundation of China (Grant No. 62202451), CAS Project for Young Scientists in Basic Research (Grant No. YSBR-029), and CAS Project for Youth Innovation Promotion Association.

\bibliographystyle{splncs04}
\bibliography{ref}
\end{document}